\documentclass{ARKkapproc}
\usepackage{epsfig}
\usepackage{subeqn}
 \kluwerbib
\newcommand{\gbf}[1] {\mbox{\boldmath${#1}$\unboldmath}}
\newcommand{\be}{\begin{equation}}
\newcommand{\ee}{\end{equation}}
\newcommand{\beq}{\begin{equation}}
\newcommand{\eeq}{\end{equation}}
\newcommand{\bed}{\begin{displaymath}}
\newcommand{\eed}{\end{displaymath}}
\newcommand{\beqa}{\begin{eqnarray}}
\newcommand{\eeqa}{\end{eqnarray}}
\newcommand{\beqann}{\begin{eqnarray*}}
\newcommand{\eeqann}{\end{eqnarray*}}
\newcommand{\bseq}{\begin{subequation}}
\newcommand{\eseq}{\end{subequation}}

\newcommand{\ba}{\begin{array}}
\newcommand{\ea}{\end{array}}

\newcommand{\negr}[1]{{\bf {#1}}}

\begin{document}
\articletitle{Kinematic analysis of \goodbreak the 3-RPR parallel manipulator}
\author{D. Chablat, Ph. Wenger}
 \affil{Institut de Recherche en Communications et Cybern\'etique
de Nantes,
 1, rue de la No\"e, 44321 Nantes, France \\
\email{Damien.Chablat\symbol{64}irccyn.ec-nantes.fr,
       ~Philippe.Wenger\symbol{64}irccyn.ec-nantes.fr}}
\author{I. Bonev}
 \affil{D\'epartement de g\'enie de la production automatis\'ee \\
      \'Ecole de Technologie Sup\'erieure \\
      1100 rue Notre-Dame Ouest, Montr\'eal (Qu\'ebec) Canada H3C 1K3\\
  \email{Ilian.Bonev\symbol{64}etsmtl.ca}}
\begin{abstract}
The aim of this paper is the kinematic study of a 3-RPR planar parallel manipulator where the three fixed revolute joints are actuated. The direct and inverse kinematic problem as well as the singular configuration is characterized. On parallel singular configurations, the motion produce by the mobile platform can be compared to the Reuleaux straight-line mechanism.
\end{abstract}
\section{Introduction}
\section{Preliminaries}
A planar three-dof manipulator with three parallel RRR chains, the
object of this paper, is shown in Fig.~1. This manipulator has
been frequently studied, in particular in \cite{Merlet,Gosselin92,Bonev2005}. The actuated joint variables are the rotation of the three revolute joints located on the base, the Cartesian variables being the position vector \negr p of the operation point $P$ and the orientation $\phi$ of the platform.

 \begin{figure}[ht]
  \begin{center}
    \includegraphics[width=80mm]{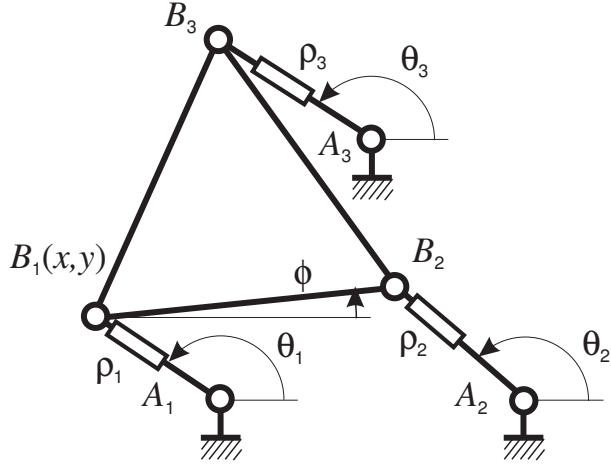}
    \caption{A three-DOF parallel manipulator}
    \protect\label{figure1}
  \end{center}
 \end{figure}
The trajectories of the points $A_i$ define an equilateral
triangle whose geometric center is the point $O$, while the points
$B_1$, $B_2$ and $B_3$, whose geometric center is the point $P$,
lie at the corners of an equilateral triangle. We thus have $||\negr a_2 - \negr a_1||=||\negr b_2 - \negr b_1||=1$, in units of length that need not be specified in the paper. .
\section{Kinematics}
The velocity $\dot{\bf p}$ of  point $P$ can be obtained in three
different forms, depending on which leg is traversed, namely,
\begin{equation}
  \dot{\negr p} = \dot{\rho}_i 
                  \frac{(\negr b_i-\negr a_i)}{||\negr b_i-\negr a_i||}
               +  \dot{\theta}_i \negr E (\negr b_i - \negr a_i)
               +  \dot{\phi} \negr E (\negr p - \negr b_i) \\
 \label{e_1}
\end{equation}
with matrix $\negr E$ defined as
 \bed
 {\bf E}= \left[\begin{array}{cr}
              0 & ~-1 \\
              1 &  0
             \end{array}
        \right]
 \eed
We would like to eliminate the three idle joint rates
$\dot{\rho}_1$, $\dot{\rho}_2$ and $\dot{\rho}_3$ from
Eqs.(\ref{e_1}), which we do upon dot-multiplying the former by
$(\negr E \negr v_i)^T$, thus obtaining
\begin{equation}
  (\negr E \negr v_i)^T \dot{\negr P}= 
  \dot{\theta_i} \rho_i +
  \dot{\phi} (\negr E \negr v_i)^T 
  \negr E (\negr p - \negr b_i) 
 \label{e_2}
\end{equation}
with 
\bed
  \negr v_i= 
  \frac{(\negr b_i-\negr a_i)}{||\negr b_i-\negr a_i||}=
  \left[
    \begin{array}{c}
     \cos(\theta_i) \\
     \sin(\theta_i)
    \end{array}
  \right] \quad {\rm and} \quad
  \rho_i= (\negr E \negr v_i)^T
          \negr E (\negr b_i - \negr a_i)
\eed
Equations (\ref{e_2}) can now be cast in vector form, namely,
\begin{equation}
  {\bf A} {\negr t}={\bf B \dot{\gbf\theta}} \quad {\rm with} \quad
  \negr t=\left[\begin{array}{c}
                 \dot{\negr p} \\
                 \dot{\phi}
            \end{array}
          \right] \quad {\rm and } \quad
  \dot{\gbf \theta}=\left[\begin{array}{c}
                 \dot{\theta}_1 \\
                 \dot{\theta}_2 \\
                 \dot{\theta}_3
            \end{array}
          \right]
  \label{e:Adp=Bdth}
\end{equation}
with $\dot{\gbf{\theta}}$ thus being the vector of actuated joint
rates.

Moreover, \negr A and \negr B are, respectively, the
direct-kinematics and the inverse-kinematics matrices of the
manipulator, defined as
\begin{equation}
 \negr A= \left[\begin{array}{cc}
   (\negr E \negr v_1)^T & ~~
   -(\negr E \negr v_1)^T \negr E (\negr p - \negr b_1) \\
   (\negr E \negr v_2)^T & ~~
   -(\negr E \negr v_1)^T \negr E (\negr p - \negr b_2) \\
   (\negr E \negr v_3)^T & ~~
   -(\negr E \negr v_3)^T \negr E (\negr p - \negr b_3)
            \end{array}
         \right] \quad
  \negr B=
  \left[\begin{array}{ccc}
   \rho_1&
   0 &
   0 \\
   0 &
   \rho_2&
   0 \\
   0 &
   0 &
   \rho_3
        \end{array}
  \right]
 \label{equation:matrices_B}
\end{equation}
When \negr A and \negr B are nonsingular, we obtain the relations
 \bed
   \negr t = \negr J \dot{\gbf \theta} {\rm,~~with~~} \negr J = \negr A^{-1} \negr B
   \quad {\rm and} \quad
   \dot{\gbf \theta} = \negr K \negr t
 \eed
with \negr K denoting the inverse of \negr J.
\subsection{Parallel Singularities}
Parallel singularities occur when the determinant of matrix \negr
A vanishes (\cite{Chablat1998} and \cite{Gosselin1990}). At these configurations, it
is possible to move locally the operation point $P$ with the
actuators locked, the structure thus resulting cannot resist
arbitrary forces, and control is lost. To avoid any performance
deterioration, it is necessary to have a Cartesian workspace free
of parallel singularities. 

For the planar manipulator studied, the direct-kinematic matrix can be written as function of $\theta_i$
\begin{equation}
\negr A= 
 \left[
   \begin{array}{ccc}
     -\sin{\theta_1} & \cos{\theta_1} & 0 \\
     -\sin{\theta_2} & \cos{\theta_2} & \cos(\theta_2) \\
     -\sin{\theta_3} & \cos{\theta_3} &
      \cos(\theta_3)/3+\sin(\theta_3)\sqrt{3}/2 
   \end{array}
 \right]
\end{equation}
and its determinant is the follows,
\begin{eqnarray}
 \det(\negr A)&=&
-2\,\sin(\theta_3-\theta_1-\theta_2) 
-\sin(\theta_3-\theta_1+\theta_2) 
-\sin(\theta_3+\theta1-\theta_2) \nonumber\\
&+&\sqrt{3}\cos(\theta_3+\theta1-\theta_2) 
-\sqrt {3}\cos(\theta_3-\theta1+\theta_2)
\end{eqnarray}

\begin{figure}[ht]
  \begin{center}
    \includegraphics[scale=1]{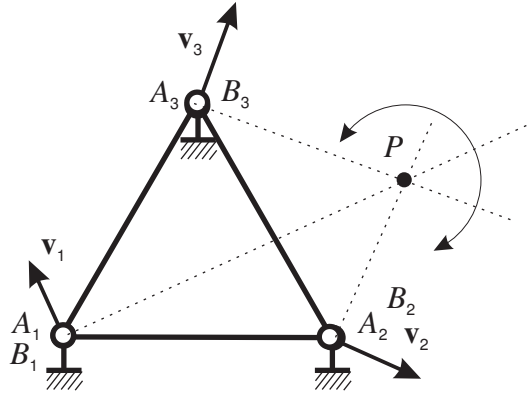}
    \caption{Parallel singularity when the three axes normal to the prismatic joint intersect}
    \protect\label{figure2}
  \end{center}
 \end{figure}

The parallel singularities occur whenever the three axes normal to the prismatic joint intersect. In such configurations, the manipulator cannot resist a wrench applies at the intersecting point $P$, as depicted in Fig.~\ref{figure2}

\begin{figure}[ht]
  \begin{center}
    \includegraphics[scale=1]{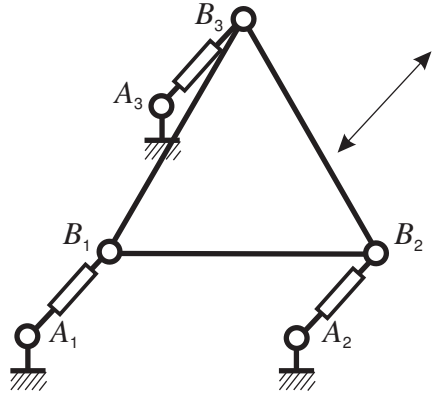}
    \caption{Parallel singularity when $\theta_1=\theta_2=\theta3$}
    \protect\label{figure_singularite_p}
  \end{center}
 \end{figure}

A special case exists when $\theta_1=\theta_2=\theta3$ because the intersection point is in infinity. Thus, the mobile platform can translate along the axes of the prismatic joints, as depicted in Fig.~\ref{figure_singularite_p}.
\subsection{Serial Singularities}
\begin{figure}[ht]
  \begin{center}
    \includegraphics[scale=1]{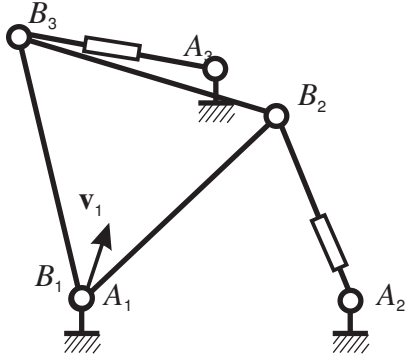}
    \caption{Serial singularity when $\rho_1=0$}    
    \protect\label{figure_singularite_s}
  \end{center}
 \end{figure}

Serial singularities occur when $\det(\negr B) = 0$. In the
presence of theses singularities, there is a direction along which
no Cartesian velocity can be produced. Serial singularities define
the boundary of the Cartesian workspace. For the topology under
study, the serial singularities occur whenever at least one $\rho_i=0$ (Figure~\ref{figure_singularite_s}). When $\rho_1=\rho_2=\rho_3=0$ not any motion can be produce by the actuated joints.
\subsection{Direct kinematics}
The position of the base joints are defined in the base reference frame,
\begin{equation}
  \negr a_1= 
  \left[
    \begin{array}{c}
    0 \\ 0 
    \end{array}
  \right] \quad
  \negr a_2=
  \left[
    \begin{array}{c}
    1 \\ 0
    \end{array}
  \right] \quad
  \negr a_3=
  \left[
    \begin{array}{c}
    1/2 \\ \sqrt{3}/2
    \end{array}
  \right]
  \label{equation:Ai}
\end{equation}
In the same way, the position of the mobile platform joints are defined in the mobile reference frame,
\begin{equation}  
  \negr b_1'=
  \left[
    \begin{array}{c}
    0 \\ 0
    \end{array}
  \right] {\rm ~}
  \negr b_2'=
  \left[
    \begin{array}{c}
    1 \\ 0 
    \end{array}
  \right] {\rm ~}
  \negr r_3'=
  \left[
    \begin{array}{c}
    1/2 \\ \sqrt{3}/2
    \end{array}
  \right]
  \label{equation:Bi}
\end{equation}
The rotation matrix \negr R describes the orientation of the mobile frame with respect to the base frame. 
\begin{eqnarray}
  \negr R= 
  \left[
    \begin{array}{cc}
     \cos(\phi) & -\sin(\phi) \\
     \sin(\phi) &  \cos(\phi)
    \end{array}
  \right] \nonumber
\end{eqnarray}
The position of $\negr p$ can be obtain in three different ways:
\begin{subequations}
\begin{eqnarray}
&&
\left\{
\begin{array}{l}
x=\cos(\theta_1) \rho_1 \\
y=\sin(\theta_1) \rho_1 
\end{array}
\right. \\
{\rm or~}&&
\left\{
\begin{array}{l}
x=-\cos(\phi)+\cos(\theta_2) \rho_2 + 1\\
y=-\sin(\phi)+\sin(\theta_2) \rho_2 \\
\end{array}
\right.\\
{\rm or~}&&
\left\{
\begin{array}{l}
x=-\cos(\phi+\pi/3)+\cos(\theta_3) \rho_3 + 1/2\\
y=-\sin(\phi+\pi/3)+\sin(\theta_3) \rho_3 + \sqrt{3}/2
\end{array}
\right.
\end{eqnarray}
\label{eq_xy}
\end{subequations}
To remove $\rho_i$ from the previous equations, we multiply $\sin(\theta_i)$ (respectively $\cos(\theta_i)$) the equations in $x$ (respectively in $y$) and we subtract the first one to the second one, to obtain three equations,
\begin{subequations}
 \begin{eqnarray}
  \sin(\theta_1)x-\cos(\theta_1)y&=&0\\
  \sin(\theta_2)x-\cos(\theta_2)y+\sin(\theta_2-\phi)-\sin(\theta_2)&=&0\\
  \sin(\theta_3)x-\cos(\theta_3)y-\cos(\theta_3-\phi+\pi/6) \nonumber \\
  -\sin(\theta_3)/2 +\cos(\theta_3) \sqrt {3}/2&=&0
 \end{eqnarray}
 \label{eq_contraintes}
\end{subequations}
We obtain $x$ and $y$ as function of $\phi$ by using Eqs.~\ref{eq_contraintes}a-b
\begin{subequations}
 \begin{eqnarray}
 x\!\!\!&=&\!\!\!
 \frac{
  \sin(\theta_2-\phi+\theta_1) 
 -\sin(\theta_1-\theta_2+\phi) 
 -\sin(\theta_1+\theta_2) 
 +\sin(\theta_1-\theta_2) }
 {2(\sin(\theta_1-\theta_2) )} \nonumber\\
 y\!\!\!&=&\!\!\!
 \frac{
  \cos(\theta_1-\theta_2+\phi) 
 -\cos(\theta_2-\phi+\theta_1) 
 -\cos(\theta_1-\theta_2) 
 +\cos(\theta_1+\theta_2)}
 {2(\sin(\theta_1-\theta_2) )} \nonumber
 \end{eqnarray}
\end{subequations}
Thus, we substitute $x$ and $y$ in Eq.~\ref{eq_contraintes}c to obtain 
\begin{subequations}
\begin{equation}
  m \cos(\phi) + n \sin(\phi) - m=0
  \label{eq_mn}
\end{equation}
with
 \begin{eqnarray}
 m&=&
   \sin(\theta_3) \cos(\theta_2) \sin(\theta_1) 
  +\sqrt {3}\,\cos(\theta_3) \sin(\theta_2) \cos(\theta_1) \nonumber \\
 &-&2\,\cos(\theta_3) \sin(\theta_2) \sin(\theta_1) 
  -\sqrt{3}\,\cos(\theta_3)\cos(\theta_2) \sin(\theta_1) \nonumber \\
 &+&\sin(\theta_3) \sin(\theta_2) \cos(\theta_1) \\
 n&=& 
  \cos(\theta_3) \sin(\theta_2) \cos(\theta_1) +
  \sqrt{3}\,\sin(\theta_3) \sin(\theta_2) \cos(\theta_1) \nonumber \\
 &-&2\,\sin(\theta_3) \cos(\theta_2) \cos(\theta_1) 
  -\sqrt{3}\,\sin(\theta_3) \cos(\theta_2) \sin(\theta_1)  \nonumber \\
 &+&\cos(\theta_3) \cos(\theta_2) \sin(\theta_1) 
 \end{eqnarray}
\end{subequations}
Equation~\ref{eq_mn} admits two roots
\begin{equation}
\phi=0 \quad  {\rm and} \quad
 \phi=\tan^{-1}
 \left( {\frac{2\,nm}{{n}^{2}+{m}^{2}}},
       -{\frac{-{m}^{2}+{n}^{2}}{{n}^{2}+{m}^{2}}} 
 \right)
\end{equation}
A trivial solution is $\phi=0$ which exists for any values of the actuated joint values where $x=y=0$. This means that when configuration of the mobile platform associated to the trivial solution to the direct kinematic problem is also a serial singularity because $\rho_1=\rho_2=\rho_3=0$.

Such a behavior is equivalent to that of the {\em agile eye}....
\subsection{Inverse kinematics}
From Eqs.~\ref{eq_contraintes}, we can easily solve the inverse kinematic problem and find two real solutions in $]-\pi~\pi]$ for each leg,
\begin{subequations}
 \begin{eqnarray}
   \theta_1&=& tan^{-1}\left(\frac{y}{x} \right) + k \pi\nonumber \\
   \theta_2&=& tan^{-1}\left({\frac {y+\sin(\phi)}
                                    {x+\cos(\phi) -1}} 
                       \right) + k \pi\nonumber \\
   \theta_3&=& tan^{-1}\left(
   {\frac {y+\sin(\phi+\pi/3) -\sqrt {3}/2}
          {x+\cos(\phi+\pi/3) -1/2}} 
                        \right) + k \pi\nonumber
 \end{eqnarray}
\end{subequations}
for $k=0, 1$ and from Eqs.~\ref{eq_xy}, we can easily find $\rho_i$,
\begin{subequations}
 \begin{eqnarray}
   \rho_1&=& \sqrt{x^2+y^2} \nonumber \\
   \rho_2&=& \sqrt{(x+\cos(\phi)-1)^2+(y+\sin(\phi))^2}\nonumber \\
   \rho_3&=& \sqrt{(x+\cos(\phi+\pi/3)-1/2)^2+
                   (y+\sin(\phi+\pi/3)-\sqrt{3}/2)^2}\nonumber
 \end{eqnarray}
\end{subequations}
\subsection{Full cycle motion: Cardanic curve}
A planar Cardanic curve is obtained by the displacement of one point of one body whose position is constrained by making two of its points lie on two coplanar lines. This curve is obtained when we set for example $\theta_1$ and $\theta_2$ and we observe the location of $B_3$. A geometrical method to solve the direct kinematic problem is find the intersection points between this curve and the line passing through the third prismatic joint whose orientation is given by $\negr v_3$. For the manipulator under study, the Cardanic curve always intersects $A_3$ and, apart from the singular configurations, $I$.

The loop $(A_1, B_1, B_2, A_2)$ can be written,
\begin{subequations}
 \begin{eqnarray}
  \rho_1 \,\cos(\theta_1) +\cos(\phi) -1- \rho_2 \,\cos(\theta_2)&=&0 \\
  \rho_1 \,\sin(\theta_1) +\sin(\phi) -\rho_2\,\sin(\theta_2)&=&0
 \end{eqnarray}
\end{subequations}
We solve the former system to have $\rho_1$ and $\rho_2$ as a function of $\phi$ and $\theta_i$ 
\begin{subequations}
 \begin{eqnarray}
\rho_1&=&{\frac 
{-\cos(\phi) \sin(\theta_2)+\sin(\theta_2)+\cos(\theta_2)\sin(\phi)}
{\cos(\theta_1)\sin(\theta_2)-\cos(\theta_2)\sin(\theta_1)}} \\
\rho_2&=&{\frac 
 {-\sin(\theta_1)\cos(\phi)+\sin(\theta_1)+\sin(\phi)\cos(\theta_1)}
 {\cos(\theta_1)\sin(\theta_2)-\cos(\theta_2)\sin(\theta_1) }}
 \end{eqnarray}
 \label{rho_i_caradanic}
\end{subequations}
The position of $B_3$ in the base frame can be written as follows,
\begin{subequations}
 \begin{eqnarray}
x_{B_3}&=&
\rho_1\,\cos(\theta_1) +\cos( \phi+\pi/3) \\
y_{B_3}&=&
\rho_1\,\sin(\theta_1) -\sin(\phi+\pi/3)
 \end{eqnarray}
 \label{eq_B_i}
\end{subequations}
$\!\!$Thus, by substituting the values of $\rho_1$ and $\rho_2$ obtained in Eqs.~\ref{rho_i_caradanic}a-b in Eqs.~\ref{eq_B_i}, we obtain the equations of the Caradanic curve,
\begin{subequations}
 \begin{eqnarray}
x_{B_3}&=&
\left({\frac
   { -\cos(\theta_1) \cos(\theta_2)}
   {\cos(\theta_2)\sin(\theta_1)-\cos(\theta_1) \sin(\theta_2)}}
   -\frac{\sqrt {3}}{2} 
\right) 
\sin( \phi)  \nonumber \\
&+&
\left( 
{\frac 
{ \cos(\theta_1)\sin(\theta_2)}
{\cos(\theta_2)\sin(\theta_1)-\cos(\theta_1)\sin(\theta_2)}}+
 \frac{1}{2}
 \right) 
\cos(\phi) \nonumber \\
&-&
\left({
{\frac { \cos(\theta_1) \sin(\theta_2)}
       {\cos(\theta_2) \sin(\theta_1)-\cos(\theta_1) \sin(\theta_2) 
}}}\right) \\
y_{B_3}&=& 
\left({\frac 
{-\cos(\theta_2) \sin(\theta_1)}
{\cos(\theta_2) \sin(\theta_1) -\cos(\theta_1) \sin(\theta_2)}}
+\frac{1}{2}\right) 
\sin(\phi) \nonumber \\
&+& \left( 
{\frac
{ \sin(\theta_1) \sin(\theta_2) }
{\cos(\theta_2) \sin(\theta_1)-\cos(\theta_1) \sin(\theta_2)}}
+\frac{\sqrt {3}}{2} \right) 
 \cos(\phi) \nonumber \\
 &-&{\frac {\sin(\theta_1) \sin(\theta_2) }
           {\cos(\theta_2) \sin(\theta_1)-\cos(\theta_1) \sin(\theta_2) }}
 \end{eqnarray}
\end{subequations}

 \begin{figure}[ht]
  \begin{center}
    \includegraphics[scale=1]{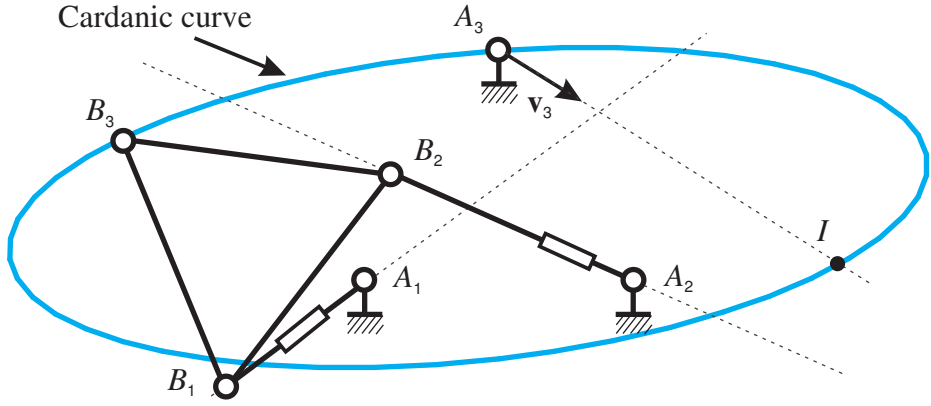}
    \caption{Cardanic curve of $B_3$ when $B_1$ and $B_2$ sliding along $\negr v_1$ and $\negr v_2$, respectively}
    \protect\label{figure_couplage}
  \end{center}
 \end{figure}
 
\begin{figure}[ht]
  \begin{center}
    \includegraphics[scale=1]{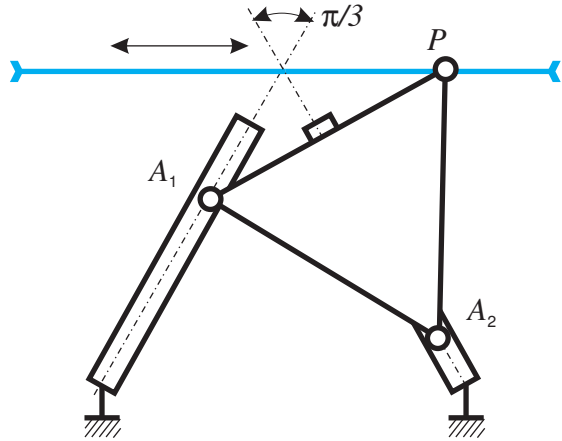}
    \caption{Reuleaux straight-line mechanism}
    \protect\label{figure_reuleaux}
  \end{center}
 \end{figure}

The Cardanic curve degenerates for specific actuated joint values (\cite{Hunt:1982}). In this case, the manipulator under study is equivalent to a {\em Reuleaux straight-line mechanism} (\cite{Nolle:1974}). This mechanism is composed by two prismatic joint and a mobile platform assembled with the prismatic joint via two revolute joints as depicted in Fig.~\ref{figure_reuleaux}. The angle between the axes passing through the prismatic joints is $\pi/3$. The mobile platform is an unit equilateral triangle. The displacement made by $P$ is a straight line whose length is two. The magnitude of this displacement is the same for $A_1$ and $A_2$.

Whenever $\theta_2-\theta_1=\pi/3$ or $\theta_1-\theta_2=\pi/3$ and $\theta_3-\theta_1=\pi/3$ or $\theta_1-\theta_3=\pi/3$ with $\theta_1\neq\theta_2\neq\theta_3$, there exists a infinity of solutions to the direct kinematic problem.
 \begin{figure}[ht]
  \begin{center}
    \includegraphics[scale=1]{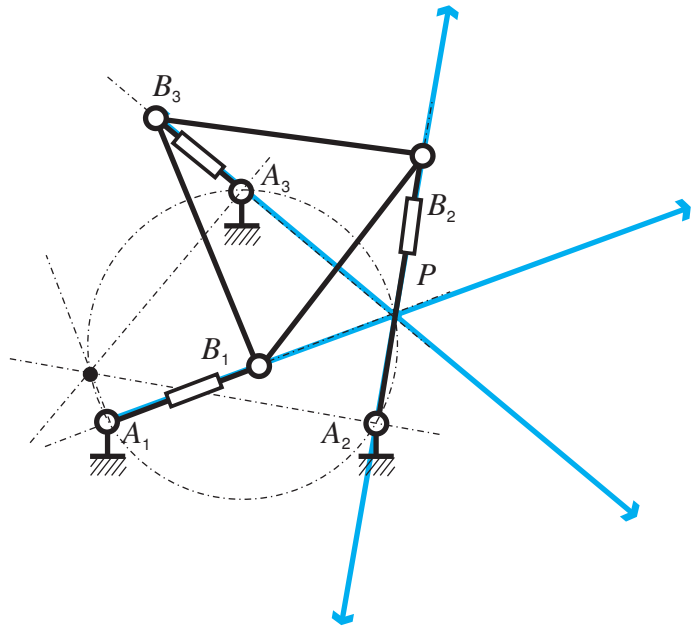}
    \caption{Degenerated Cardanic curve when $\theta_2-\theta_1=\pi/3$}
    \protect\label{figure5}
  \end{center}
 \end{figure}
This feature exists whenever we have the same condition to have a parallel singularity and the three vector $\negr v_i$ intersect in one point. The displacement of $A_i$ are around this point $P$ and is magnitude is $4 \sqrt{3}/3$.
\section{Conclusions}
A kinematic analysis of a planar 3-RPR parallel manipulator was presented in this paper. The parallel and serial singularities have been characterized as well as the direct and inverse kinematics. This mechanism features two direct kinematic solutions whose one is a trivial singular configuration. For some actuated joint values associated to a parallel singularity, the motion made by the mobile platform is equivalent to a Reuleaux straight-line mechanism where the amplitude of the motion is well known.


\begin{chapthebibliography}{1}
\bibitem[Gosselin, 1990]{Gosselin1990}
Gosselin, C. and Angeles, J., {\em Singularity Analysis of Closed-Loop Kinematic Chains}, IEEE, Transaction on Robotics and Automation, Vol. 6, pp. 281-290, June 1990.
\bibitem[Chablat, 1998]{Chablat1998}
Chablat, D. and Wenger, Ph., {\em Working Modes and Aspects in
Fully-Parallel Manipulator}, Proceedings IEEE International Conference on Robotics and Automation, 1998, pp. 1964-1969.
\bibitem[Merlet, 2000]{Merlet}
Merlet, J-P., ``Parallel robots,'' Kluwer Academic Publ., Dordrecht, The Netherland, 2000.
\bibitem[Gosselin, 1992]{Gosselin92}
Gosselin, C., Sefrioui, J., Richard, M. J., {\em Solutions polynomiales au probl\`eme de la cin\'ematique des manipulateurs parall\`eles plans \`a trois degr\'e de libert\'e}, Mechanism and Machine Theory, Vol. 27, pp. 107-119, 1992.
\bibitem[Bonev, 2005]{Bonev2005}
Bonev, I. and Gosselin C.M., ``Singularity Loci of Spherical Parallel Mechanisms'', Proceedings IEEE International Conference on Robotics and Automation, 2005.
\bibitem[Tischler, 1998]{Tischler1998}
Tischler C.R., Hunt K.H., Samuel A.E., ``A Spatial Extension of Cardanic Movement: its Geometry and some Derived Mechanisms'', Mechanism and MachineTheory, No.~33, pp.~1249--1276, 1998.
\bibitem[Nolle, 1974]{Nolle:1974}
Nolle H., ``Linkage coupler curve synthesis: A historical review—II. Developments after 1875'', Mechanism and Machine Theory, Vol.~9, Issues 3-4, pp.~325--348, 1974
\bibitem[Hunt, 1982]{Hunt:1982}
Hunt K.H. and Herman P.M., ``Multiple Cusps, Duality and Cardanic Configurations in Planar Motion'', Mechanism and Machine Theory, Vol.~17, No.~6, pp.~361--368, 1982
\end{chapthebibliography}
\end{document}